\documentclass[runningheads]{llncs}
\usepackage{graphicx}
\usepackage{url}
\usepackage{todonotes}
\usepackage{xcolor}
\usepackage{colortbl}
\usepackage{listings}
\usepackage{xspace}
\usepackage{svg}
\usepackage{arydshln}
\usepackage{hyperref}
\usepackage{enumitem}

\newcommand{\ExeKGLib}{\texttt{ExeKGLib}\xspace}

\newcommand{\owlindiv}{\texttt{owl:Individual}\xspace}
\newcommand{\owlindivs}{\texttt{owl:Individuals}\xspace}

\newcommand{\footnotehref}[1]{\footnote{\href{#1}{#1}}}
\newcommand{\grayline}{\arrayrulecolor{gray}\hline\arrayrulecolor{black}}

\makeatletter
\AtBeginDocument{%
  \def\doi#1{\url{https://doi.org/#1}}}
\makeatother

\begin{document}
\title{ExeKGLib: Knowledge Graphs-Empowered Machine Learning Analytics}
%
%
\author{Antonis Klironomos\inst{1,2} \and
Baifan Zhou\inst{3} \and
Zhipeng Tan\inst{1,4} \and
Zhuoxun Zheng\inst{1,5} \and
Gad-Elrab Mohamed\inst{1} \and
Heiko Paulheim\inst{2} \and
Evgeny Kharlamov\inst{1,3}}
\authorrunning{Klironomos et al.}
%
\institute{
Bosch Center for Artificial Intelligence, Germany \and
University of Mannheim, Germany \and
University of Oslo, Norway \and
RWTH Aachen, Germany \and
Oslo Metropolitan University, Norway
}
\maketitle              
\begin{abstract}
Many machine learning (ML) libraries are accessible online for ML practitioners. Typical ML pipelines are complex and consist of a series of steps, each of them invoking several ML libraries. In this demo paper, we present \ExeKGLib, a Python library that allows users with coding skills and minimal ML knowledge to build ML pipelines. \ExeKGLib relies on knowledge graphs to improve the transparency and reusability of the built ML workflows, and to ensure that they are executable. We demonstrate the usage of \ExeKGLib and compare it with conventional ML code to show \ExeKGLib's benefits.\looseness=-1

\end{abstract}

\section{Introduction}
\vspace{-1ex}
Due to the significant advancements in the realm of computer science, particularly in the field of machine learning (ML), there is a plethora of ML algorithms and corresponding libraries publicly accessible~\cite{obulesu2018machine,Bartschat19,mikut2017matlab,heidrich2021pywatts}. The use of ML is steadily rising in both academic and industrial settings~\cite{sarkerMachineLearningAlgorithms2021}. Experts in various domains are also learning ML for the sake of applying it to solve domain-specific challenges,  e.g. biologists \cite{libbrechtMachineLearningApplications2015,kimInfectiousDiseaseOutbreak2021}, oncologists \cite{kourouMachineLearningApplications2015,abreuPredictingBreastCancer2016}, and engineers in the industry \cite{mengMachineLearningAdditive2020,zeng2020importance,huang2023hybrid}. The development of functional and useful ML workflows can be complex and time-consuming, which can pose a barrier for non-ML experts. Thus, there is a need for a user-friendly approach that neither requires excessive knowledge nor training in ML. While, existing tools such as Amazon Sage Maker\footnotehref{https://aws.amazon.com/sagemaker} or Google AutoML\footnotehref{https://cloud.google.com/automl} provide convenient graphical user interfaces (GUI) and application programming interfaces (API), yet do not provide open-source code libraries. 

In this paper, we introduce \ExeKGLib, an easily-extendable Python library that supports a variety of methods for data visualization, data preprocessing and feature engineering, and ML modeling. \ExeKGLib works in two steps: (1) Generate executable ML pipelines using knowledge graphs (KGs), (2) Convert generated pipelines into functional Python scripts, and execute these scripts. We rely on KGs for expressing the created pipelines to make them more understandable and reusable, and to verify that they are executable \cite{zhengExecutableKnowledgeGraphs2022a}.
\ExeKGLib can be used by a wide range of users and in a variety of scenarios: from domain experts that want to do ML to teachers and students for teaching and learning ML.

\begin{figure}[t]
    \vspace{-2ex}
    \centering
    \includegraphics[width=\textwidth]{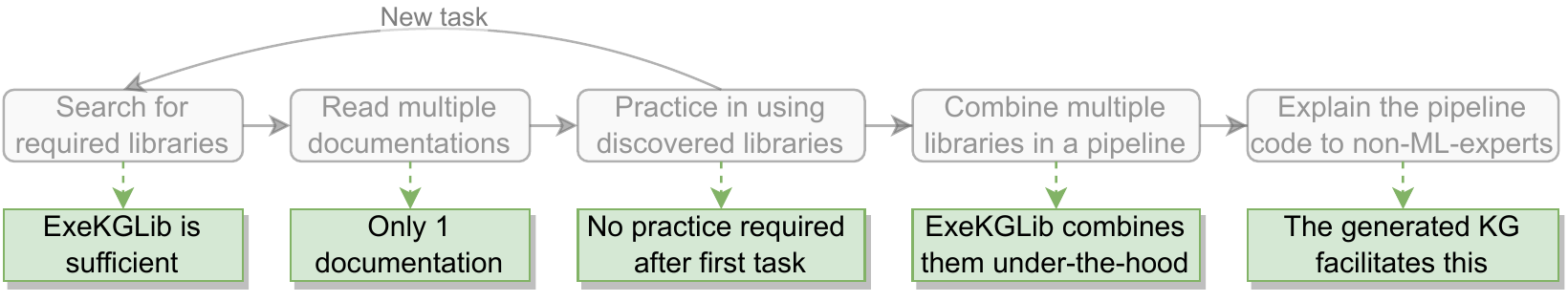}
    \vspace{-6ex}
    \caption{Improvements on conventional data science workflow} \label{fig:ds-workflow-comparison}
    \vspace{-4ex}
\end{figure}

In the following sections, we start with demonstrating \ExeKGLib's usage. Then, we describe the used KG schemata and discuss the underlying details of KG construction and pipeline generation in Section~\ref{system-design}.

\section{Usage Demonstration}

Our target user can generate an ML pipeline either by importing \ExeKGLib's \texttt{ExeKG} Python module or by interacting with the provided Typer CLI without writing code~\footnotehref{https://github.com/boschresearch/ExeKGLib\#usage}. We demonstrate the former usage with three sample Python files~\footnotehref{https://github.com/boschresearch/ExeKGLib/tree/main/examples}. The pipelines represented by the generated sample KGs are briefly explained below:

\begin{enumerate}
    \item \textbf{ML pipeline}: Loads features and labels from an input CSV dataset, splits the data, trains and tests a k-NN model, and visualizes the prediction errors.
    \item \textbf{Statistics pipeline}: Loads a feature from an input CSV dataset, normalizes it, and plots its values (before and after normalization) using a scatter plot.
    \item \textbf{Visualization pipeline}: Loads a feature from an input CSV dataset and plots its values using a line plot.
\end{enumerate}

The above pipelines (in form of executable KGs) can be executed using the provided Typer CLI~\footnotehref{https://github.com/boschresearch/ExeKGLib\#executing-an-ml-pipeline}. To exhibit the pipelines' transparency, we have visualized the sample pipelines using Neo4j~\footnotehref{https://bit.ly/exe-kg-lib-visualizations}. The script to perform this visualization for any executable KG is also provided.

Experimentation with the offered resources can verify the benefits of \ExeKGLib on the traditional data science process (Figure \ref{fig:ds-workflow-comparison}). In particular, using our tool to solve a task reduces the overhead prior to the implementation, reduces the effort during the code development, and increases the explainability of the resulting ML pipeline. A brief display of the tool's practical advantages for a generic classification task is illustrated in Table \ref{comparison-of-code}.
In a conventional setting (table's middle column), the user needs to separately import three different libraries (\textit{i.e.} \texttt{pandas}, \texttt{scikit-learn}, \texttt{matplotlib}) and use five of their modules. On the other hand, when using \ExeKGLib (table's right column), the user needs a limited number of libraries and modules, and thus learning is easier and faster by skipping reading extensive documentation of various libraries.\looseness=-1

\section{System Design}
\label{system-design}

\begin{table}[t]
\vspace{-1ex}
    \centering
    \caption{Comparison between conventional code and \ExeKGLib for a classification task}\label{comparison-of-code}
    \vspace{-2ex}
    \begin{tabular}{|l|l|l|l|}
        \hline
        \textbf{Pipeline steps} & \textbf{Conventional code} & \textbf{Code using ExeKGLib} \\
        \hline
        1. Load data & \texttt{pd.read\_csv()} + convert to \texttt{numpy} & \begin{tabular}{@{}l@{}}\texttt{ExeKG.create\_data\_entity()} \\ \texttt{ExeKG.create\_pipeline\_task()}\end{tabular} \\
        \grayline
        2. Split data & \texttt{sklearn}...\texttt{train\_test\_split()} & \texttt{ExeKG.add\_task()} \\
        \grayline
        3. Train & \texttt{sklearn}...\texttt{Classifier().fit()} & \texttt{ExeKG.add\_task()} \\
        \grayline
        4. Evaluate & \texttt{sklearn}...\texttt{Classifier().predict()} & \texttt{ExeKG.add\_task()} \\
        \grayline
        5. Visualize & \texttt{matplotlib.pyplot}...\texttt{()} & \texttt{ExeKG.add\_task()} \\
        \hline
    \end{tabular}
    \vspace{-2ex}
\end{table}

\ExeKGLib relies on KG schemata to construct executable KGs (representing an ML pipeline) and execute them. Both of these processes use the \texttt{rdflib} Python library combined with SPARQL queries to find and create KG components.\looseness=-1


\vspace{-1ex}
\subsection{Underlying KG Schemata}
\label{kg-schemata}
\vspace{-1ex}

\ExeKGLib utilizes an upper-level KG schema (Data Science -- namespace: \texttt{ds}) that describes data science concepts such as data entity, task, and method. The supported tasks and methods are separated into bottom-level KG schemata~\footnotehref{https://github.com/boschresearch/ExeKGLib\#kg-schemata}: 

\begin{itemize}[topsep=3pt,parsep=0pt,partopsep=0pt,itemsep=0pt,leftmargin=*]
\item \textbf{Visualization} tasks schema, which includes two types of methods:
(1) The plot canvas methods that define the plot size and layout. (2) The various kinds of plot methods (e.g. line plot, scatter plot, or bar plot). 
\item \textbf{Statistics and Feature Engineering} tasks schema including methods such as Interquartile Range calculation, mean
and standard deviation calculation, etc., which can also form more complex
methods like outlier detection and normalization.
\item \textbf{ML} tasks schema representing ML algorithms like Linear Regression, MLP, and k-NN and helper functions that perform e.g. data splitting and ML model performance calculation.
\end{itemize}

\ExeKGLib's Python implementations of the above methods utilize common libraries such as \texttt{matplotlib} and \texttt{scikit-learn}.

\vspace{-1ex}
\subsection{Executable KG Construction}
\label{kg-construction}
\vspace{-1ex}

\begin{figure}[t]
    \vspace{-4ex}
    \centering
    \includegraphics[width=.9\textwidth]{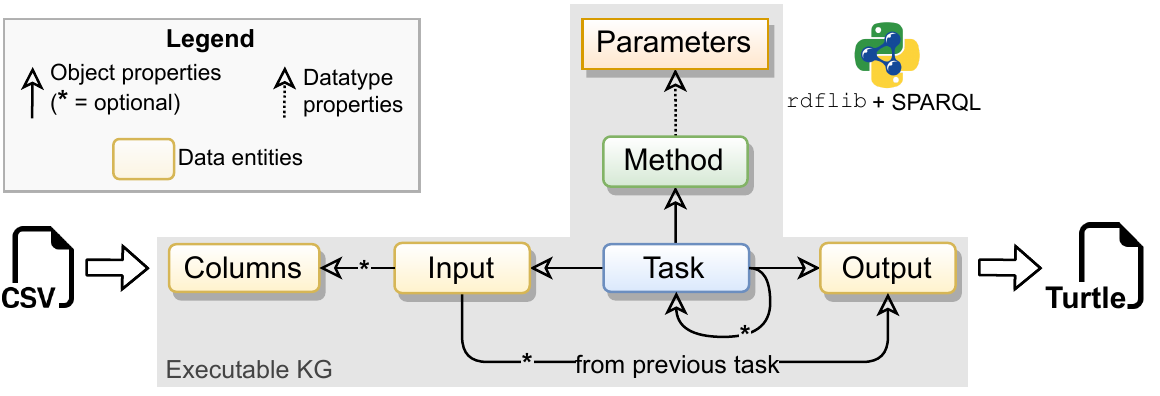}
    \vspace{-3ex}
    \caption{Executable KG construction phase} \label{fig:kg-construction}
    \vspace{-3ex}
\end{figure}

As shown in Figure \ref{fig:kg-construction}, the internal process of creating an executable KG starts with extracting the columns from the input dataset (CSV file). \ExeKGLib populates the KG with data entities representing the target columns. Data entities are then used as input to the ML pipeline tasks.

Afterward, \ExeKGLib adds to the KG the entities representing the user-specified task type (e.g. classification) and method type (e.g. k-NN), which are taken from the provided bottom-level KG schemata; and links the current task with the chosen method, input data entities, datatype properties, and the next task. Throughout the process, the compatibility of the aforementioned KG components is ensured by \ExeKGLib based on the KG schemata. Finally, the created KG is serialized and saved on the disk in Turtle.

\subsection{ML Pipeline Execution}
\label{ml-pipeline-execution}
To execute a given KG, \ExeKGLib parses the KG with the help of the above KG schemata (Section~\ref{kg-schemata}). After that, the pipeline's \textit{Tasks} (\owlindivs) are sequentially traversed using the object property \texttt{ds:hasNextTask}. Based on the IRI of the next \textit{Task} (\owlindiv), the \textit{Task}'s type and properties are retrieved and mapped dynamically to a Python object. Such mapping allows for extending the library without modifying the KG execution code. Finally, for each \textit{Task}, the Python implementation of the selected method type is invoked.
\vspace{-2ex}
\section{Future Work}

We plan to add additional algorithms to \ExeKGLib to support a wider variety of ML-related tasks, which can be conveniently done due to its good extendability. In the future, we will build a system by integrating \ExeKGLib with a graph-based database. This will allow for easier management of the produced executable KGs, quick visualization, and more convenient reuse.

\noindent \subsubsection*{Acknowledgements}:
The work was partially supported by EU projects Dome 4.0 (GA 953163), OntoCommons (GA 958371), DataCloud (GA 101016835), Graph Massiviser (GA 101093202), and EnRichMyData (GA 101093202).

\bibliographystyle{splncs04}
\bibliography{references}

\begin{thebibliography}{10}
\providecommand{\url}[1]{\texttt{#1}}
\providecommand{\urlprefix}{URL }
\providecommand{\doi}[1]{https://doi.org/#1}

\bibitem{abreuPredictingBreastCancer2016}
Abreu, P.H., Santos, M.S., Abreu, M.H., Andrade, B., Silva, D.C.: Predicting
  {{Breast Cancer Recurrence Using Machine Learning Techniques}}: {{A
  Systematic Review}}. ACM Computing Surveys  \textbf{49}(3),  52:1--52:40 (Oct
  2016). \doi{10.1145/2988544}

\bibitem{Bartschat19}
Bartschat, A., Reischl, M., Mikut, R.: {Data Mining Tools}. Wiley
  Interdisciplinary Reviews: Data Mining and Knowledge Discovery
  \textbf{9}(4),  e1309 (2019). \doi{10.1002/widm.1309}

\bibitem{heidrich2021pywatts}
Heidrich, B., Bartschat, A., Turowski, M., Neumann, O., Phipps, K.,
  Meisenbacher, S., Schmieder, K., Ludwig, N., Mikut, R., Hagenmeyer, V.:
  {pyWATTS: Python Workflow Automation Tool for Time Series}. arXiv preprint
  arXiv:2106.10157  (2021). \doi{10.48550/arXiv.2106.10157}

\bibitem{huang2023hybrid}
Huang, Z., Fey, M., Liu, C., Beysel, E., Xu, X., Brecher, C.: {Hybrid
  Learning-Based Digital Twin for Manufacturing Process: Modeling Framework and
  Implementation}. Robotics and Computer-Integrated Manufacturing  \textbf{82},
   102545 (2023). \doi{10.1016/j.rcim.2023.102545}

\bibitem{kimInfectiousDiseaseOutbreak2021}
Kim, J., Ahn, I.: {Infectious Disease Outbreak Prediction Using Media Articles
  with Machine Learning Models}. Scientific Reports  \textbf{11}(1), ~4413 (Feb
  2021). \doi{10.1038/s41598-021-83926-2}

\bibitem{kourouMachineLearningApplications2015}
Kourou, K., Exarchos, T.P., Exarchos, K.P., Karamouzis, M.V., Fotiadis, D.I.:
  {Machine Learning Applications in Cancer Prognosis and Prediction}.
  Computational and Structural Biotechnology Journal  \textbf{13},  8--17 (Jan
  2015). \doi{10.1016/j.csbj.2014.11.005}

\bibitem{libbrechtMachineLearningApplications2015}
Libbrecht, M.W., Noble, W.S.: {Machine Learning Applications in Genetics and
  Genomics}. Nature Reviews Genetics  \textbf{16}(6),  321--332 (Jun 2015).
  \doi{10.1038/nrg3920}

\bibitem{mengMachineLearningAdditive2020}
Meng, L., McWilliams, B., Jarosinski, W., Park, H.Y., Jung, Y.G., Lee, J.,
  Zhang, J.: Machine {{Learning}} in {{Additive Manufacturing}}: {{A Review}}.
  JOM  \textbf{72}(6),  2363--2377 (Jun 2020). \doi{10.1007/s11837-020-04155-y}

\bibitem{mikut2017matlab}
Mikut, R., Bartschat, A., Doneit, W., Ordiano, J.{\'A}.G., Schott, B.,
  Stegmaier, J., Waczowicz, S., Reischl, M.: {The MATLAB Toolbox SciXMiner:
  User’s Manual and Programmer’s Guide}. arXiv preprint arXiv:1704.03298
  (2017). \doi{10.48550/arXiv.1704.03298}

\bibitem{obulesu2018machine}
Obulesu, O., Mahendra, M., ThrilokReddy, M.: {Machine Learning Techniques and
  Tools: A Survey}. In: 2018 International Conference on Inventive Research in
  Computing Applications (ICIRCA). pp. 605--611. IEEE (2018).
  \doi{10.1109/ICIRCA.2018.8597302}

\bibitem{sarkerMachineLearningAlgorithms2021}
Sarker, I.H.: Machine {{Learning}}: {{Algorithms}}, {{Real-World Applications}}
  and {{Research Directions}}. SN Computer Science  \textbf{2}(3), ~160 (Mar
  2021). \doi{10.1007/s42979-021-00592-x}

\bibitem{zeng2020importance}
Zeng, L., Al-Rifai, M., Chelaru, S., Nolting, M., Nejdl, W.: {On the Importance
  of Contextual Information for Building Reliable Automated Driver
  Identification Systems}. In: 2020 IEEE 23rd International Conference on
  Intelligent Transportation Systems (ITSC). pp.~1--8. IEEE (2020).
  \doi{10.1109/ITSC45102.2020.9294439}

\bibitem{zhengExecutableKnowledgeGraphs2022a}
Zheng, Z., Zhou, B., Zhou, D., Zheng, X., Cheng, G., Soylu, A., Kharlamov, E.:
  Executable {{Knowledge Graphs}} for {{Machine Learning}}: {{A Bosch Case}} of
  {{Welding Monitoring}}. In: The {{Semantic Web}} \textendash{} {{ISWC}} 2022,
  vol. 13489, pp. 791--809. {Springer International Publishing}, {Cham} (2022).
  \doi{10.1007/978-3-031-19433-7_45}

\end{thebibliography}
\end{document}